\documentclass{article}
\usepackage{ijcai24}

\usepackage{times}
\usepackage{soul}
\usepackage{url}
\usepackage[hidelinks]{hyperref}
\usepackage[utf8]{inputenc}
\usepackage[small]{caption}
\usepackage{graphicx}
\usepackage{amsmath}
\usepackage{amsthm}
\usepackage{amssymb}
\usepackage{booktabs}
\usepackage{algorithm}
\usepackage{bbm}
\usepackage{algorithmic}
\usepackage[switch]{lineno}

\title{Multi-Modality Spatio-Temporal Forecasting via Self-Supervised Learning}

\author{
Jiewen Deng$^1$
\and
Renhe Jiang$^{2}$\thanks{Corresponding author.}
\and
Jiaqi Zhang$^1$\And
Xuan Song$^{1,3*}$
\affiliations
$^1$Southern University of Science and Technology\\
$^2$The University of Tokyo\\
$^3$Jilin University
\emails
dengjw1@outlook.com,
jiangrh@csis.u-tokyo.ac.jp\\
zhangjq0@outlook.com,
songx@sustech.edu.cn
}

\begin{document}

\maketitle

\begin{abstract}
Multi-modality spatio-temporal (MoST) data extends spatio-temporal (ST) data by incorporating multiple modalities,  which is prevalent in monitoring systems, encompassing diverse traffic demands and air quality assessments. Despite significant strides in ST modeling in recent years, there remains a need to emphasize harnessing the potential of information from different modalities. Robust MoST forecasting is more challenging because it possesses (i) high-dimensional and complex internal structures and (ii) dynamic heterogeneity caused by temporal, spatial, and modality variations. In this study, we propose a novel \underline{Mo}ST learning framework via \underline{S}elf-\underline{S}upervised \underline{L}earning, namely \textbf{MoSSL}, 
which aims to uncover latent patterns from temporal, spatial, and modality perspectives while quantifying dynamic heterogeneity.
Experiment results on two real-world MoST datasets verify the superiority of our approach compared with the state-of-the-art baselines. Model implementation is available at \url{https://github.com/beginner-sketch/MoSSL}.
\end{abstract}

\section{Introduction}
\begin{figure}[t]
  \centering
  \includegraphics[width=.95\linewidth]{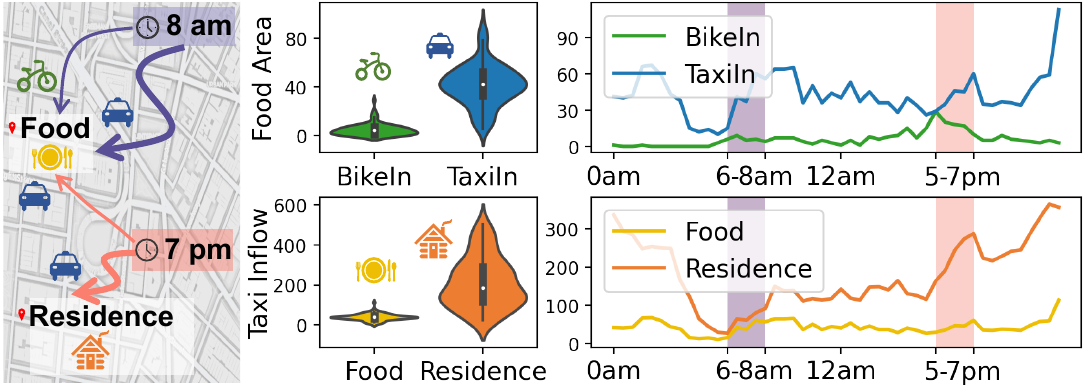}
  \caption{Illustration of heterogeneous NYC traffic demand data. Heterogeneity exists in the modalities (Bike Inflow and Taxi Inflow), locations (Food and Residence), and time periods (8 am and 7 pm).}
  \label{fig:intro}
\end{figure}

Multi-modality spatio-temporal (MoST) data is ubiquitous in real-world applications. Robust MoST prediction is crucial for various monitoring systems in smart city, particularly for sensitive scenarios that require subsequent decision-making. Compared to spatio-temporal (ST) data, MoST integrates information from an additional modality domain beyond the spatio-temporal domains. For example, urban public transportation management involves data from different time periods, locations, and transportation modalities, i.e., $\{$Bike, Taxi$\}\times$$\{$Inflow, Outflow$\}$~\cite{zhang2023automated,ijcai2023p231}. Although MoST data enables a more comprehensive and diverse feature acquisition, it is non-trivial to model them due to its significant heterogeneity and dynamics. Specifically, modeling MoST data presents two critical challenges. The first challenge lies in \textbf{precisely obtaining variations and correlations} of different modalities across time and space. While using one modality to supervise others \cite{huang2019mist,wu2020hierarchically} is common practice, this results in late-fusion models ignoring inter-modality correlations or, in early-fusion models, a trade-off between inter-modality and intra-modality features. The second challenge lies in \textbf{accurately quantifying heterogeneous components} to protect the model from erroneous pattern variations. Taking NYC traffic demand data as an example, in Figure~\ref{fig:intro}, we depict observations of two modalities (i.e., Bike/Taxi Inflow) across different locations over a period and the corresponding distributions. Findings indicate that both modalities exhibit an increase in the Food area at 8 am, yet Taxi Inflow significantly surpasses Bike Inflow; by 7 pm, Taxi Inflow peaks in the Residential zone while remaining steady in the Food area. Such heterogeneities across space, time, and modality amplify uncertainties and errors in MoST modeling. 

For more effective time series modeling, the existing methods have made significant strides by modeling latent spatial correlations among sensors and temporal auto-correlation within time series. Mainstream models utilize GCN-based \cite{wu2020connecting,shao2022decoupled} modules for spatial relations and sequence models such as RNN \cite{lai2018modeling}, TCN \cite{wu2020connecting}, and attention mechanisms \cite{wu2021autoformer,zheng2020gman} for temporal relations. These techniques focus more on lower-level hierarchical information and overlook valuable insights about spatio-temporal relationships provided by multiple modalities. 
A new line of research increasingly extracts evolution from other modalities as auxiliary inputs, adding supplemental information to leverage similar historical patterns \cite{fang2021mdtp,han2021dynamic}. However, these modality-aware practices predominantly emphasize \textbf{benefiting spatio-temporal aspects}, posing a challenge in preserving interactions within the high-dimensional latent space in MoST due to the parameter space shared among modalities. 
Recently, motivated by the success of self-supervised learning ~\cite{hendrycks2019using}, especially contrastive learning achieved in text, image, and audio domain ~\cite{giorgietal2021declutr}, researchers apply SSL principles to model time series data~\cite{10.1145/3550316,ji2023spatio,gao2023spatio}. 
But these approaches primarily concentrate on \textbf{modeling domain-specific heterogeneity}, be it spatio-temporal or modality-specific, without fully encompassing multiple domains. MoST modeling requires comprehensive consideration of interactions and heterogeneity across space, time, and modality to obtain a more thorough and accurate representation.
Thus far, despite systematic exploration of spatio-temporal regularities, modality interactions are adequately unaddressed, let alone signals entangled in MoST. 

%Therefore, in this study, to address/capture/xxx/, we propose a

Different modalities of spatio-temporal data record the observations from different perspectives, promising to improve the forecasting performance by mining the latent patterns and dynamic heterogeneity deeply and sufficiently. Therefore, we are motivated to propose a novel \underline{Mo}ST learning framework via \underline{S}elf-\underline{S}upervised \underline{L}earning, namely \textbf{MoSSL}. It comprises the following steps: (i) implementing a MoST Encoder to model the information from the space, time, and modality; (ii) introducing a Multi-modality Data Augmentation to understand pattern correlations governed by modality rules and integrate MoST domain information; (iii) devising a Global Self-Supervised Learning (GSSL) to discern diverse pattern changes among different perspectives; (iv) developing a Modality Self-Supervised Learning (MSSL) to further strengthen learning representations of inter-modality and intra-modality features in high-dimensional feature space. The contributions of our work are summarized as follows:
\begin{itemize}
    \item We propose a novel MoST learning framework, called MoSSL, to explicitly uncover latent patterns and disentangle the heterogeneity from temporal, spatial, and modality perspectives.
    \item We design a Multi-modality Data Augmentation, coupled with two auxiliary self-supervised paradigms, i.e., GSSL and MSSL, which empower the model to discern and quantify dynamic heterogeneity in MoST.
    \item We conduct extensive experiments on two real-world MoST datasets (i.e., the traffic dataset with four transportation modalities and the air quality dataset with three pollutants). The results demonstrate that our model outperforms multiple state-of-the-art baselines.
\end{itemize}

\section{Preliminaries}
In this section, we formally describe the multi-modality spatio-temporal forecasting problem.

\noindent \textbf{Problem Statement.} The time series information of all spatial units (i.e., nodes) $N$ and modalities $M$ at $t^{\text{th}}$ time slot is denoted as $X_t \in \mathbb{R}^{N \times M}$. Given $T$ consecutive time steps, the multi-modality spatio-temporal forecasting problem aims to forecast the next $O$ steps as $\hat{Y} \in \mathbb{R}^{O \times N\times M}$, which can be expressed by the following conditional distribution:
\begin{equation}
P(\hat{Y}|X) = \prod\nolimits_{t=1}^{O} P(\hat{Y}_{t,:,:}|X)
\end{equation}

\begin{figure*}[t]
  \centering
  \includegraphics[width=\linewidth]{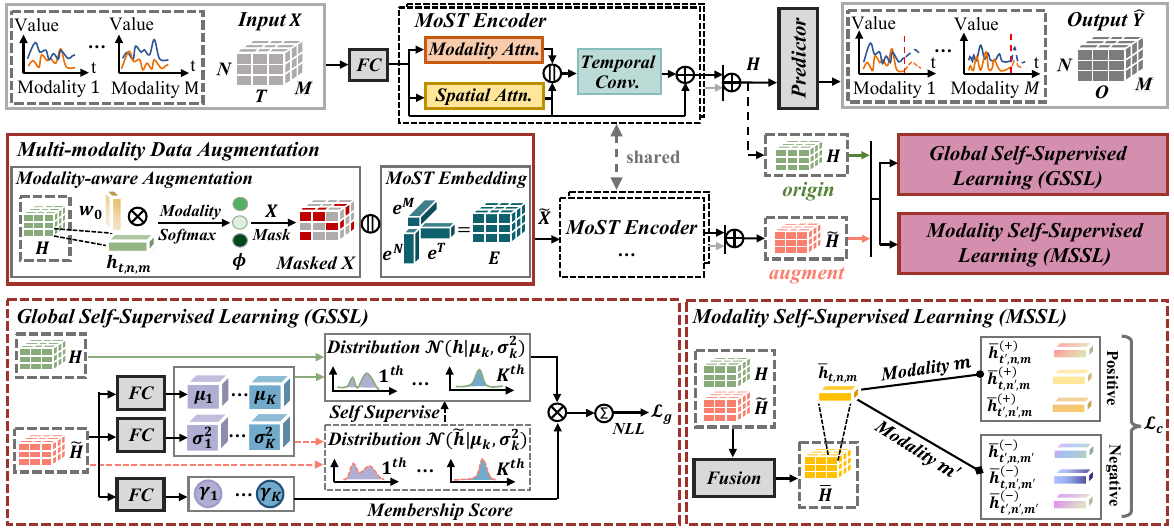}
  \caption{The proposed \textbf{MoSSL} framework: (i) Takes a three-dimensional MoST $X$ as input, generating the original representation $H$ through a multi-layered MoST Encoder; (ii) Simultaneously, Multi-modality Data Augmentation refines $X$ into $\Tilde{X}$, feeding it into the shared MoST Encoder to produce the augmented representation $\Tilde{H}$; (iii) With $H$ and $\Tilde{H}$ available, deploys Global Self-Supervised Learning (GSSL) and Modality Self-Supervised Learning (MSSL) to generate losses $\mathcal{L}_g$ and $\mathcal{L}_c$.}
  \label{fig:framework}
\end{figure*}

\section{Methodology}
In this section, we will introduce the technical details of the proposed MoST learning framework, namely \textbf{MoSSL}, as demonstrated in Figure~\ref{fig:framework}.

% \subsection{MoST Encoder}
\subsection{Multi-modality Spatio-Temporal Encoder}
We design a Multi-modality Spatio-Temporal (MoST) Encoder that collectively preserves context in MoST data, allowing the modeling of sequential patterns across different time periods, geographical correlations between spatial nodes, and shared knowledge among modalities. For convenience, we define a non-linear projection function with ReLU activation function as $f(x)=relu(x \cdot w+b)$, where $w$ and $b$ are learnable parameters. Specifically, the MoST Encoder takes $H^{(\text{in})}=f(X) \in \mathbb{R}^{T \times N \times M \times d_z}$ as input.

\noindent \textbf{Modality Attention.} Inspired by \cite{zheng2020gman}, we design a Modality Attention (MA) based on the attention mechanism to encode the modality-wise correlations. For modality $m$ at time step $t$ and node $n$, MA works as follows:
\begin{equation}
\begin{aligned}
u_{m,m^{\prime}} &= \frac{\left< f_1^{(\text{ma})}(h_{t,n,m}^{(\text{in})}), f_2^{(\text{ma})}(h_{t,n,m^{\prime}}^{(\text{in})})  \right>}{\sqrt{d_z}}\\
\alpha_{m,m^{\prime}} &= \frac{exp(u_{m,m^{\prime}})}{\sum_{i \in M} exp(u_{m,i})}\\
h_{t,n,m}^{(\text{ma})} &= \sum\nolimits_{m^{\prime} \in M} \alpha_{m,m^{\prime}} \cdot f_3^{(\text{ma})}(h_{t,n,m^{\prime}}^{(\text{in})})
\end{aligned}
\end{equation}
\noindent where $\left< \cdot \right>$ denotes the inner product operator, $u_{m,m^{\prime}}$ indicates the relevance between modality $m$ and $m^{\prime}$, and $\alpha_{m,m^{\prime}}$ is the attention score that indicates the importance of modality $m^{\prime}$ to $m$. Thus, MA can adaptively capture inter-modality correlations and generate weighted modality representation $H^{(\text{ma})} \in \mathbb{R}^{T \times N \times M \times d_z}$.

\noindent \textbf{Spatial Attention.} We employ a Spatial Attention (SA), akin to MA, to model dynamic spatial relationships among nodes.
For node $n$ at time step $t$ in modality $m$, the spatial correlation representation $H^{(\text{sa})} \in \mathbb{R}^{T \times N \times M \times d_z}$ can be calculated by:
\begin{equation}
\begin{aligned}
s_{n,n^{\prime}} &= \frac{\left< f_1^{(\text{sa})}(h_{t,n,m}^{(\text{in})}), f_2^{(\text{sa})}(h_{t,n^{\prime},m}^{(\text{in})})  \right>}{\sqrt{d_z}}\\
\beta_{n,n^{\prime}} &= \frac{exp(s_{n,n^{\prime}})}{\sum_{i \in N} exp(s_{n,i})}\\
h_{t,n,m}^{(\text{sa})} &= \sum\nolimits_{n^{\prime} \in N} \beta_{n,n^{\prime}} \cdot f_3^{(\text{sa})}(h_{t,n^{\prime},m}^{(\text{in})})
\end{aligned}
\end{equation}
\noindent where $s_{n,n^{\prime}}$ indicates the relevance between node $n$ and $n^{\prime}$, and $\beta_{n,n^{\prime}}$ is the attention score that indicates the importance of node $n^{\prime}$ to $n$.

\noindent \textbf{Temporal Convolution.} We implement a Temporal Convolution (TC) with a tangent filter and a sigmoid gate~\cite{Oord2016WaveNetAG} to extract high-level temporal features. 
Formally, given the representation $\hat{H}=[H^{(\text{in})},H^{(\text{ma})},H^{(\text{sa})}] \in \mathbb{R}^{T \times N \times M \times 3d_z}$, it works as follows:
\begin{equation}
\resizebox{.9\linewidth}{!}{$
\displaystyle
H^{(\text{tc})} = f_{\text{conv}}(\text{tanh}(\hat{H} \ast W_{\text{filter}}) \odot \sigma (\hat{H} \ast W_{\text{gate}}))
$}
\end{equation}
\noindent where $\ast$ denotes the dilated causal convolution, $W_{\text{filter}}$ and $W_{\text{gate}}$ are learnable weights, and $\odot$ indicates element-wise multiplication. $f_{\text{conv}}(\cdot)$ is a 1$\times$1 convolution function.

We construct our MoST Encoder with a multi-layer structure, incorporating MA and SA in conjunction with TC. This architecture serves as the foundation for representing the spatio-temporal-modality relationships and yields the original representation $H \in \mathbb{R}^{T \times N \times M \times d_z}$ as its output.
\subsection{Multi-modality Data Augmentation}
% \subsection{MoST Data Augmentation}
Here, we introduce two data augmentation strategies,i.e., Modality-aware Augmentation and MoST Embedding.

\noindent \textbf{Modality-aware Augmentation.}
We devise a Modality-aware Augmentation operator to provide tailored modality dependencies for each instance. Different from the common method for spatial graph augmentation \cite{jiang2023spatio,ji2023spatio}, our operator learns the heterogeneity ruled by modalities based on the original representation $H$:
\begin{equation}
\resizebox{.91\linewidth}{!}{$
\displaystyle
\phi_{t,n,m} = \frac{exp(v_{t,n,m})}{\sum\limits_{i \in M} exp(v_{t,n,i})}, \text{where} \hspace{3pt} v_{t,n,m} = h_{t,n,m} \cdot w_0
$}
\end{equation}
\noindent where $\phi_{t,n,m}$ is the aggregation weight indicating the modality relevance, $w_0 \in \mathbb{R}^{d_z}$ is a learnable vector. With $\phi_{t,n,m}$ available, we can derive the mask probability from the Bernoulli distribution, i.e., $p_{t,n,m} \sim Bern(1-\phi_{t,n,m})$, and conduct the augmentation on the input tensor $X$. Higher $p_{t,n,m}$ signifies stronger modality heterogeneity (lower correlation), increasing the likelihood of masking the data $x_{t,n,m}$.

\noindent \textbf{MoST Embedding.}
Since the evolving patterns of MoST are confined by space, time, and modality constraints, we develop an adaptive MoST Embedding to incorporate available domain information into the model. Specifically, we create spatial and modality embeddings to encode respective features into $e_n \in \mathbb{R}^{d_z}$ and $e_m \in \mathbb{R}^{d_z}$, capturing static information. To enrich dynamic correlations, we generate temporal embedding to encode each historical time step into $e_t \in \mathbb{R}^{d_z}$. Then, we obtain the adaptive MoST embedding $e_{t,n,m}=e_t+e_n+e_m$, denoted as $E \in \mathbb{R}^{T \times N \times M \times d_z}$.

After obtaining two augmentations, we concatenate them as $\Tilde{X}$, and transform it by using a non-linear projection $\Tilde{H}^{(\text{in})}=f(\Tilde{X})$, which is fed into the shared MoST Encoder to generate the augmented representation $\Tilde{H} \in \mathbb{R}^{T \times N \times M \times d_z}$.

\subsection{Global Self-Supervised Learning}
% \subsection{MoST Self-Supervised Learning}
Recently, cluster-based learning has disrupted the independence between samples by leveraging latent clustering information. By tapping into additional prior knowledge derived from clustering, the aim is to retain higher-level heterogeneous information within the learned representations. However, existing time series modeling methods rely on clustering partitions that might not necessarily indicate the optimal solution. Additionally, some approaches \cite{zhu2021mixseq,duan2022plae,ji2023spatio} assume temporal constancy between representations and cluster memberships, which are not applicable for intricate and dynamic MoST data.
We introduce a novel Global Self-Supervised Learning (GSSL) to address the above limitations, which could explicitly capture the heterogeneous patterns across spatial, temporal, and modality domains. Furthermore, we expect to preserve the dynamic attributes based on the underlying context, showcasing both independence and complementarity. For instance, some clusters correspond to modality properties, remaining consistent across space over time, while others correspond to temporal identities, possibly having different change frequencies over time. 

Specifically, with an augmented representation $\Tilde{H}$ derived from an augmented input $\Tilde{X}$, GSSL seeks to model a mixture of Gaussian distributions over $\Tilde{h}_{t,n,m}$ within $\Tilde{H}$. Technically, the probability distribution is conditioned on the augmented input, but we omit this condition in the subsequent formula for simplicity.
Formally, given the augmented representation $\Tilde{H}$, we first infer the probability distribution of $\Tilde{h}_{t,n,m} \in \mathbb{R}^{d_z}$, which can be decomposed as follows:
\begin{equation}
P(\Tilde{h}_{t,n,m})=\sum\nolimits_{k=1}^K \gamma_k(\Tilde{h}_{t,n,m}) P(\Tilde{h}_{t,n,m}| \gamma_k(\Tilde{h}_{t,n,m}))
\label{eq:cluster}
\end{equation}
where $K$ is the number of clusters (heterogeneous components). $\gamma_k(\Tilde{h}_{t,n,m})$ indicates the membership score associated with $\Tilde{h}_{t,n,m}$ belonging to the $k^\text{th}$ cluster, satisfying $0 \leq \gamma_k(\Tilde{h}_{t,n,m}) \leq 1$ and $\sum\nolimits_k \gamma_k(\Tilde{h}_{t,n,m}) =1$. Then, we assign $\gamma_k(\Tilde{h}_{t,n,m})$ via a linear mapping parameterized with learnable weight $W^{\gamma}_{k} \in \mathbb{R}^{T L S d_z}$:
\begin{equation} 
\gamma_k(\Tilde{h}_{t,n,m}) = \frac{\exp(W^{\gamma}_{k} \cdot \;\text{vec}(\Tilde{H}))}{\sum_{j=1}^K \exp(W^{\gamma}_{j} \cdot \;\text{vec}(\Tilde{H}))}
\end{equation}
Without loss of generality, we assume $P(\Tilde{h}_{t,n,m}| \gamma_k(\Tilde{h}_{t,n,m}))$ follows a Gaussian distribution $\mathcal{N}(\Tilde{h}_{t,n,m}|\mu_{k},\sigma_{k}^2)$, and the mean $\mu_{k} \in \mathbb{R}^{d_z}$ and variance $\sigma^2_{k} \in \mathbb{R}^{d_z}$ are generated by:
\begin{equation}
\begin{aligned}
\mu_{k,d} &= W^{\mu}_k\cdot\text{vec}(\Tilde{H}[..., d]) +b^{\mu}_{k}\\
\sigma^2_{k,d} &= \exp\left(W^{\sigma}_{k} \cdot \text{vec}\left(\Tilde{H}[..., d]\right) +b^{\sigma}_{k}\right)
\end{aligned}
\label{eq:para}
\end{equation}
where $W_{k,d}^{\mu}, W^{\sigma}_{k,d} \in \mathbb{R}^{ T L S}$, $b^{\mu}_{k,d}, b^{\sigma}_{k,d} \in \mathbb{R}$ are learnable weights and biases, $d \in [1,...,d_z]$ is the hidden channel index. $\exp(\cdot)$ ensures that $\sigma^2_{k}$ remains greater than zero.

After obtaining the distribution of the augmented representation, i.e., $\mathcal{N}(\Tilde{h}_{t,n,m}|\mu_{k},\sigma_{k}^2)$, we treat it as a self-supervised signal. It aims to align the distribution of the original representation $h_{t,n,m}$ with that of $\Tilde{h}_{t,n,m}$, performed by $\mathcal{N}(h_{t,n,m}|\mu_{k},\sigma_{k}^2)$. This practice leverages data augmentation to enrich the original representation $h$, enhancing its understanding of heterogeneous components. From the probability perspective, we replace $P(\Tilde{h}_{t,n,m}| \gamma_k(\Tilde{h}_{t,n,m}))$ (in Eq.~\ref{eq:cluster}) with $P(h_{t,n,m}| \gamma_k(\Tilde{h}_{t,n,m}))$, and apply the Negative Log-Likelihood (NLL) to optimize the self-supervised task:
\begin{equation}
\resizebox{.91\linewidth}{!}{$
\displaystyle
\mathcal{L}_g = - \sum_{t,n,m}^{T,N,M} \log \left\{ \sum_{k=1}^K \gamma_k(\Tilde{h}_{t,n,m}) P(h_{t,n,m}| \gamma_k(\Tilde{h}_{t,n,m})) \right\}
$}
\label{eq:cluloss}
\end{equation}

By aligning the distributions of $H$ and $\Tilde{H}$, the original representation $H$ can reflect the global heterogeneity across time, space, and modality, aiding the model forecasting.

\subsection{Modality Self-Supervised Learning}
% \subsection{Cross-Modality Self-Supervised Learning}
Recent self-supervised contrastive learning methods often compare representations from two distorted versions of the same input instance (i.e., original and augmented views) \cite{grill2020bootstrap} or across historical time intervals \cite{ji2023spatio}. Differing from these methods, we devised a Modality Self-Supervised Learning (MSSL) paradigm to further enhance the model's understanding of cross-modality heterogeneity, mitigating the loss of information due to modality ignorance or time sensitivity. This brings two advantages: (i) aiding the model in better utilizing modality information and finer learning of spatio-temporal features across different modalities; (ii) reducing interference among different modalities, enabling a more precise capture of the spatio-temporal dependencies within each modality.

Specifically, given the representations $H$ and $\Tilde{H}$ from the original and augmented views, we first fuse them with learnable weights $w_1$, $w_2 \in \mathbb{R}^{d_z}$, and provide the unified modality representation $c_{\mathcal{M}}$ through aggregating spatio-temporal features as follows:
\begin{equation}
\begin{aligned}
r_{t,n,m} &= h_{t,n,m}\odot w_1 + \Tilde{h}_{t,n,m} \odot w_2\\
c_{\mathcal{M}} &= \sigma (\frac{1}{TN}\sum_{t,n}^{T,N} r_{t,n,m})
\end{aligned}
\end{equation}

Then, the representation $\bar{h}_{t,n,m}=(r_{t,n,m}, c_{\mathcal{M}})$ is used for positive and negative pairs sampling. Within our sampling strategy, the modality-aligned instances (intra-modality) are considered positive pairs, and the instances across all available modalities (inter-modality) are considered negative pairs. Here, the idea is to obtain the modality-specific patterns and encourage the discovery of heterogeneity across different modalities. Following the convention, we derive the objective by Binary Cross Entropy (BCE) with a sigmoid function for numerical stability:
\begin{equation}
\resizebox{.89\linewidth}{!}{$
\displaystyle
\mathcal{L}_c = - \sum_{t,n,m}^{T,N,M} \left\{ \log\mathcal{G}(\bar{h}_{t,n,m}) + \log(1-\mathcal{G}(\bar{h}_{t,n,m^{\prime}})) \right\}
$}
\label{eq:clloss}
\end{equation}
where $\mathcal{G}(\bar{h}_{t,n,m^{\prime}})=\sigma(r_{t,n,m^{\prime}}^\top w_3 c_{\mathcal{M}})$ is the transforming function, $m$ and $m^{\prime}$ indicate two different modalities, and $w_3 \in \mathbb{R}^{d_z \times d_z}$ is the learnable parameter shared by positive and negative pairs.

\subsection{Model Optimization}
For the MoST forecasting task, our predictor comprises two fully connected layers with ReLU as the activation function, adapting the dimension of the hidden channel to the desired output dimension. It takes the representation $H$ from the original view as input:
\begin{equation}
\begin{aligned}
\hat{Y} = relu(relu(H) \cdot W_{\text{o1}}+b_{\text{o1}}) \cdot W_{\text{o2}} + b_{\text{o2}}
\end{aligned}
\end{equation}
\noindent where $W_{\text{o1}} \in \mathbb{R}^{d_k \times d_k}$, $W_{\text{o2}} \in \mathbb{R}^{d_k \times O}$, $b_{\text{o1}} \in \mathbb{R}^{d_k}$, and $b_{\text{o2}} \in \mathbb{R}^{O}$ are learnable weights and biases. $\text{relu}(\cdot)$ denotes ReLU function that is applied element-wise. $\hat{Y} \in \mathbb{R}^{O \times N\times M}$ indicates the forecasting result of the next $O$ steps for $N$ nodes and $M$ modalities.

In the learning phase, we incorporate two self-supervised learning losses (i.e., Eq.~\ref{eq:cluloss} and Eq.~\ref{eq:clloss}) into the objective criterion (i.e., MSE) as the overall optimization:
\begin{equation}
\mathcal{L} = \underbrace{\sum_{t,n,m}^{O,N,M} {\parallel Y_{t,n,m}-\hat{Y}_{t,n,m} \parallel}^2}_{\mathcal{L}_r} + \mathcal{L}_g + \mathcal{L}_c
\label{eq:mse}
\end{equation}

\section{Experiments}
In this section, to evaluate the performance of our MoSSL, we develop a group of research questions (RQ) and conduct a series of experiments correspondingly:

\noindent \textbf{RQ1:} How does MoSSL perform compared with the state-of-the-art baselines? (Section \ref{overall_performance})

\noindent \textbf{RQ2:} How does MoSSL perform compared with its model variants? (Section \ref{ablation_study})

\noindent \textbf{RQ3:} How is the interpretability of MoSSL? (Section \ref{case_study})

\noindent \textbf{RQ4:} What are the effects of hyper-parameters in MoSSL? (Section \ref{hyper})

\subsection{Experiment Setup}
\begin{table}[h]
    \scriptsize
    % \tiny
    \centering
    \setlength{\tabcolsep}{1mm}{
    \begin{tabular}{c||c|c}
        \hline
        \multicolumn{1}{c||}{Dataset}  & 
        \multicolumn{1}{c|}{\textbf{NYC Traffic Demand}} &
        \multicolumn{1}{c}{\textbf{BJ Air Quality}}
        \\
        \hline
        Time period &
        \multicolumn{1}{c|}{2016/4/1$\sim$2016/6/30 (0.5 hour)} &
        \multicolumn{1}{c}{2013/3/1$\sim$2017/2/28 (1 hour)}
        \\
        \hline
        % Time slot &
        % \multicolumn{1}{c|}{1 hour} &
        % \multicolumn{1}{c}{1 hour} 
        % \\
        % \hline
        Nodes &
        \multicolumn{1}{c|}{98 nodes} &
        \multicolumn{1}{c}{10 nodes}
        \\
        \hline
        Modalities & \{Bike, Taxi\}$\times$\{Inflow, Outflow\} & \{PM2.5, PM10, SO$_2$\} 
        \\
        \hline
        Horizons & \multicolumn{2}{c}{input: 16 $\rightarrow$ output: 1 / 2 / 3}
        \\
        \hline
    \end{tabular}}
    \caption{Summary of experimental datasets.}
    \label{tab:datasets}
\end{table}
% NYC traffic dataset
\begin{table*}[t]
\scriptsize
\centering
\setlength{\tabcolsep}{0.5mm}{
\begin{tabular*}{17cm}{@{\extracolsep{\fill}}cc|ccc|ccc|ccc|ccc}
	\hline
	\multicolumn{1}{c|}{Modality}  & 
	\multicolumn{2}{c|}{Bike Inflow} &
	\multicolumn{2}{c|}{Bike Outflow} &
	\multicolumn{2}{c|}{Taxi Inflow} & 
	\multicolumn{2}{c}{Taxi Outflow}
	\\
	\hline
	\multicolumn{1}{c|}{Method} &
	\multicolumn{1}{c|}{MAE} & 
	\multicolumn{1}{c|}{RMSE} &
	\multicolumn{1}{c|}{MAE} & 
	\multicolumn{1}{c|}{RMSE} &
	\multicolumn{1}{c|}{MAE} & 
	\multicolumn{1}{c|}{RMSE} &
	\multicolumn{1}{c|}{MAE} & 
	\multicolumn{1}{c}{RMSE}  
	\\
	\hline  
	\multicolumn{1}{c|}{LSTNet} &
	\multicolumn{1}{c|}{3.21 / 3.32 / 3.66} &
	\multicolumn{1}{c|}{8.65 / 9.05 / 9.72} &
	\multicolumn{1}{c|}{3.45 / 3.69 / 3.84} &
	\multicolumn{1}{c|}{9.40 / 10.14 / 10.43} &
	\multicolumn{1}{c|}{9.77 / 10.45 / 11.50} &
	\multicolumn{1}{c|}{20.66 / 21.75 / 23.81} &
	\multicolumn{1}{c|}{10.44 / 11.11 / 11.87} &
	\multicolumn{1}{c}{19.98 / 21.46 / 22.77}
	\\
	\multicolumn{1}{c|}{AGCRN} &
	\multicolumn{1}{c|}{2.40 / 2.70 / 3.02} &
	\multicolumn{1}{c|}{6.96 / 7.69 / 8.41} &
	\multicolumn{1}{c|}{2.60 / 2.97 / 3.31} &
	\multicolumn{1}{c|}{7.39 / 8.32 / 9.15} &
	\multicolumn{1}{c|}{7.08 / 8.10 / 9.16} &
	\multicolumn{1}{c|}{13.42 / 15.71 / 18.16} &
	\multicolumn{1}{c|}{7.81 / 9.03 / 10.19} &
	\multicolumn{1}{c}{15.03 / 17.41 / 19.29}
	\\
    \multicolumn{1}{c|}{MTGNN} &
	\multicolumn{1}{c|}{2.28 / 2.44 / 2.69} &
	\multicolumn{1}{c|}{6.68 / 7.05 / 7.62} &
	\multicolumn{1}{c|}{2.45 / 2.60 / 2.83} &
	\multicolumn{1}{c|}{7.07 / 7.46 / 8.02} &
	\multicolumn{1}{c|}{7.24 / 8.04 / 8.78} &
	\multicolumn{1}{c|}{13.82 / 15.82 / 17.71} &
	\multicolumn{1}{c|}{7.97 / 8.97 / 9.96} &
	\multicolumn{1}{c}{15.47 / 17.59 / 19.26}
	\\
	\multicolumn{1}{c|}{GWN} &
	\multicolumn{1}{c|}{2.37 / 2.66 / 2.91} &
	\multicolumn{1}{c|}{6.89 / 7.63 / 8.23} &
	\multicolumn{1}{c|}{2.57 / 2.91 / 3.17} &
	\multicolumn{1}{c|}{7.42 / 8.26 / 8.85} &
	\multicolumn{1}{c|}{7.00 / 8.23 / 9.59} &
	\multicolumn{1}{c|}{13.30 / 16.11 / 18.99} &
	\multicolumn{1}{c|}{7.72 / 9.05 / 10.29} &
	\multicolumn{1}{c}{14.98 / 17.48 / 19.69}
	\\
	% \multicolumn{1}{c|}{StemGNN} &
	% \multicolumn{1}{c|}{2.50 / 2.86 / 3.27} &
	% \multicolumn{1}{c|}{7.17 / 8.04 / 9.01} &
	% \multicolumn{1}{c|}{2.63 / 3.06 / 3.48} &
	% \multicolumn{1}{c|}{7.55 / 8.63 / 9.65} &
	% \multicolumn{1}{c|}{7.80 / 9.38 / 10.84} &
	% \multicolumn{1}{c|}{14.76 / 18.01 / 21.12} &
	% \multicolumn{1}{c|}{8.25 / 9.82 / 11.32} &
	% \multicolumn{1}{c}{15.83 / 18.89 / 21.62}
	% \\
	\multicolumn{1}{c|}{ST-Norm} &
	\multicolumn{1}{c|}{2.37 / 2.64 / 2.91} &
	\multicolumn{1}{c|}{6.93 / 7.54 / 8.13} &
	\multicolumn{1}{c|}{2.57 / 2.92 / 3.20} &
	\multicolumn{1}{c|}{7.42 / 8.29 / 8.95} &
	\multicolumn{1}{c|}{7.13 / 8.35 / 9.65} &
	\multicolumn{1}{c|}{13.60 / 16.28 / 18.72} &
	\multicolumn{1}{c|}{7.88 / 9.17 / 10.44} &
	\multicolumn{1}{c}{15.25 / 17.71 / 19.76}
	\\
        \multicolumn{1}{c|}{STtrans} &
	\multicolumn{1}{c|}{2.72 / 3.06 / 3.38} &
	\multicolumn{1}{c|}{7.74 / 8.54 / 9.35} &
	\multicolumn{1}{c|}{2.96 / 3.27 / 3.56} &
	\multicolumn{1}{c|}{8.33 / 8.93 / 9.67} &
	\multicolumn{1}{c|}{8.46 / 9.62 / 11.03} &
	\multicolumn{1}{c|}{15.67 / 17.74 / 20.42} &
	\multicolumn{1}{c|}{8.78 / 10.04 / 11.28} &
	\multicolumn{1}{c}{16.66 / 19.12 / 21.39}
	\\
	\multicolumn{1}{c|}{MiST} &
	\multicolumn{1}{c|}{2.32 / 2.52 / 2.76} &
	\multicolumn{1}{c|}{6.78 / 7.26 / 7.84} &
	\multicolumn{1}{c|}{2.57 / 2.83 / 3.01} &
	\multicolumn{1}{c|}{7.48 / 8.10 / 8.54} &
	\multicolumn{1}{c|}{7.93 / 8.73 / 9.61} &
	\multicolumn{1}{c|}{14.10 / 16.51 / 19.06} &
	\multicolumn{1}{c|}{8.44 / 9.19 / 10.82} &
	\multicolumn{1}{c}{15.56 / 17.93 / 19.33}
	\\
        \multicolumn{1}{c|}{COCOA} &
	\multicolumn{1}{c|}{2.82 / 3.11 / 3.48} &
	\multicolumn{1}{c|}{5.06 / 5.82 / 6.53} &
	\multicolumn{1}{c|}{2.93 / 3.28 / 3.60} &
	\multicolumn{1}{c|}{5.43 / 6.23 / 6.85} &
	\multicolumn{1}{c|}{8.05 / 9.81 / 11.95} &
	\multicolumn{1}{c|}{13.98 / 17.85 / 22.18} &
	\multicolumn{1}{c|}{8.80 / 10.69 / 12.83} &
	\multicolumn{1}{c}{16.20 / 20.15 / 24.02}
	\\
        \hline
	\multicolumn{1}{c|}{MoSSL} &
	\multicolumn{1}{c|}{\textbf{2.25} / \textbf{2.39} / \textbf{2.58}} &
	\multicolumn{1}{c|}{\textbf{4.10} / \textbf{4.28} / \textbf{4.51}} &
	\multicolumn{1}{c|}{\textbf{2.43} / \textbf{2.55} / \textbf{2.67}} &
	\multicolumn{1}{c|}{\textbf{4.42} / \textbf{4.65} / \textbf{4.86}} &
	\multicolumn{1}{c|}{\textbf{6.73} / \textbf{7.53} / \textbf{8.38}} &
	\multicolumn{1}{c|}{\textbf{11.89} / \textbf{14.13} / \textbf{16.22}} &
	\multicolumn{1}{c|}{\textbf{7.37} / \textbf{8.18} / \textbf{8.90}} &
	\multicolumn{1}{c}{\textbf{13.80} / \textbf{15.61} / \textbf{17.23}}
	\\
	\hline
\end{tabular*}}
\caption{Performance on NYC Traffic Demand Dataset for the first/second/third horizon.}
\label{tab:traf}
\end{table*}
% BJ Air Quality dataset
\begin{table*}[t]
\scriptsize
\centering
\setlength{\tabcolsep}{0.5mm}{
\begin{tabular*}{15cm}{@{\extracolsep{\fill}}cc|ccc|ccc|ccc}
	\hline
	\multicolumn{1}{c|}{Modality}  & 
	\multicolumn{2}{c|}{PM2.5} &
	\multicolumn{2}{c|}{PM10} &
	\multicolumn{2}{c}{SO$_2$} 
	\\
	\hline
	\multicolumn{1}{c|}{Method} &
	\multicolumn{1}{c|}{MAE} & 
	\multicolumn{1}{c|}{RMSE} &
	\multicolumn{1}{c|}{MAE} & 
	\multicolumn{1}{c|}{RMSE} &
	\multicolumn{1}{c|}{MAE} & 
	\multicolumn{1}{c}{RMSE}
	\\
	\hline  
	\multicolumn{1}{c|}{LSTNet} &
	\multicolumn{1}{c|}{12.98 / 17.55 / 21.76} &
	\multicolumn{1}{c|}{24.09 /  31.56 / 40.11} &
	\multicolumn{1}{c|}{19.23 / 24.93 / 30.26} &
	\multicolumn{1}{c|}{31.70 / 39.96 / 52.25} &
	\multicolumn{1}{c|}{2.15 /  3.69 / 4.47} &
	\multicolumn{1}{c}{7.13 /  10.07 / 11.41} 
	\\
	\multicolumn{1}{c|}{AGCRN} &
	\multicolumn{1}{c|}{9.85 / 15.39 / 19.83} &
	\multicolumn{1}{c|}{18.74 / 28.15 / 35.06} &
	\multicolumn{1}{c|}{15.62 / 22.37 / 27.32} &
	\multicolumn{1}{c|}{26.41 / 36.13 / 43.13} &
	\multicolumn{1}{c|}{\textbf{1.57} / 2.38 / 3.00} &
	\multicolumn{1}{c}{6.06 / 7.81 / 9.09} 
	\\
        \multicolumn{1}{c|}{MTGNN} &
	\multicolumn{1}{c|}{9.82 / 14.95 / 19.26} &
	\multicolumn{1}{c|}{18.41 / 27.35 / 34.10} &
	\multicolumn{1}{c|}{15.83 / 22.70 / 27.61} &
	\multicolumn{1}{c|}{26.30 / 36.13 / 43.14} &
	\multicolumn{1}{c|}{1.73 / \textbf{2.31} / 2.82} &
	\multicolumn{1}{c}{6.39 / 7.68 / 8.80} 
	\\
	\multicolumn{1}{c|}{GWN} &
	\multicolumn{1}{c|}{9.78 / 15.39 / 19.72} &
	\multicolumn{1}{c|}{18.57 / 28.08 / 35.10} &
	\multicolumn{1}{c|}{15.50 / 22.39 / 27.43} &
	\multicolumn{1}{c|}{26.09 / 36.22 / 43.76} &
	\multicolumn{1}{c|}{1.61 / 2.45 / 2.95} &
	\multicolumn{1}{c}{6.11 / 7.89 / 9.06} 
	\\
	% \multicolumn{1}{c|}{StemGNN} &
	% \multicolumn{1}{c|}{10.42 / 16.30 / 21.25} &
	% \multicolumn{1}{c|}{19.53 / 29.30 / 36.71} &
	% \multicolumn{1}{c|}{16.48 / 23.57 / 28.86} &
	% \multicolumn{1}{c|}{27.71 / 37.84 / 45.29} &
	% \multicolumn{1}{c|}{2.47 / 3.13 / 3.51} &
	% \multicolumn{1}{c}{8.11 / 9.40 / 10.23} 
	% \\
	\multicolumn{1}{c|}{ST-Norm} &
	\multicolumn{1}{c|}{13.16 / 17.66 / 22.14} &
	\multicolumn{1}{c|}{21.42 / 29.98 / 36.89} &
	\multicolumn{1}{c|}{18.11 / 24.06 / 28.97} &
	\multicolumn{1}{c|}{28.02 / 37.57 / 44.79} &
	\multicolumn{1}{c|}{3.47 / 3.42 / 4.32} &
	\multicolumn{1}{c}{9.04 / 9.37 / 10.87} 
	\\
        \multicolumn{1}{c|}{STtrans} &
	\multicolumn{1}{c|}{10.55 / 15.46 / 19.56} &
	\multicolumn{1}{c|}{21.88 / 31.37 / 38.71} &
	\multicolumn{1}{c|}{16.61 / 23.14 / 28.21} &
	\multicolumn{1}{c|}{30.54 / 42.60 / 50.85} &
	\multicolumn{1}{c|}{2.34 / 3.13 / 3.79} &
	\multicolumn{1}{c}{7.55 / 9.41 / 10.71} 
	\\
	\multicolumn{1}{c|}{MiST} &
	\multicolumn{1}{c|}{13.98 / 17.77 / 22.07} &
	\multicolumn{1}{c|}{21.66 / 30.77 / 35.95} &
	\multicolumn{1}{c|}{18.71 / 24.12 / 29.48} &
	\multicolumn{1}{c|}{28.20 / 38.44 / 46.42} &
	\multicolumn{1}{c|}{2.45 / 3.23 / 3.78} &
	\multicolumn{1}{c}{7.95  / 9.75 / 10.90} 
	\\
        \multicolumn{1}{c|}{COCOA} &
	\multicolumn{1}{c|}{10.53 / 15.51 / 19.44} &
	\multicolumn{1}{c|}{18.68 / 27.70 / 34.49} &
	\multicolumn{1}{c|}{16.33 / 22.88 / 27.81} &
	\multicolumn{1}{c|}{27.65 / 35.87 / 42.95} &
	\multicolumn{1}{c|}{2.89 / 3.11 / 3.84} &
	\multicolumn{1}{c}{5.94 / \underline{7.17} / 9.18} 
	\\
        \hline
	\multicolumn{1}{c|}{MoSSL} &
	\multicolumn{1}{c|}{\textbf{9.35} / \textbf{14.91} / \textbf{19.05}} &
	\multicolumn{1}{c|}{\textbf{18.32} / \textbf{27.28} / \textbf{33.85}} &
	\multicolumn{1}{c|}{\textbf{14.86} / \textbf{21.81} / \textbf{26.90}} &
	\multicolumn{1}{c|}{\textbf{25.90} / \textbf{35.94} / \textbf{42.37}} &
	\multicolumn{1}{c|}{1.63 / 3.02 / \textbf{2.62}} &
	\multicolumn{1}{c}{\textbf{5.59} / \textbf{6.25} / \textbf{7.32}} 
	\\
	\hline
\end{tabular*}}
\caption{Performance on BJ Air Quality Dataset for the first/second/third horizon.}
\label{tab:air}
\end{table*}

\noindent \textbf{Datasets.} We conduct experiments on two real-world MoST datasets, namely \textbf{NYC Traffic Demand} and \textbf{BJ Air Quality}, as listed in Table~\ref{tab:datasets}. NYC Traffic Demand dataset is collected from the New York City, which consists of 98 nodes and four transportation modalities: Bike Inflow, Bike Outflow, Taxi Inflow, and Taxi Outflow. BJ Air Quality dataset is collected from the Beijing Municipal Environmental Monitoring Center, which contains 10 nodes and three pollutant modalities: PM2.5, PM10, and SO$_2$.

\noindent \textbf{Settings.}
We implement the network with the Pytorch toolkit. For the model, the layers of MoST Encoder is 4, where the kernel size of each dilated causal convolution component is 2, and the related expansion rate is \{2, 4, 8, 16\} in each layer. This enables our model to handle the 16 input steps. The number of cluster components $K$ and the dimension of hidden channels $d_z$ are set to 4 and 48. The training phase is performed using the Adam optimizer, and the batch size is 16. In addition, the inputs are normalized by Z-Score.

\noindent \textbf{Baselines.} To evaluate our model's prediction accuracy quantitatively, We implement eight baselines that fall into two categories for comparison. 
\textbf{Non-MoST forecasting models:} 1) LSTNet \cite{lai2018modeling} is a classical time series prediction model; 2) MTGNN \cite{wu2020connecting}, 3) Graph Wavenet (GWN) \cite{wu2019graph}, and 4) AGCRN \cite{bai2020adaptive}, the most representative deep models for spatio-temporal forecasting, embed adaptive graph leaning into temporal convolution; 5) ST-Norm \cite{deng2021st} is a normalization-based approach that refines the temporal and spatial components. 
\textbf{MoST forecasting models:} 6) STtrans \cite{wu2020hierarchically}, a transformer-based model that integrates spatial, temporal and semantic dependencies; 7) MiST \cite{huang2019mist} is a classic co-predictive model that incorporates the spatio-temporal and categorical embeddings; 8) COCOA \cite{10.1145/3550316} is a modality-matching contrastive framework that minimizes the similarity between irrelevant modalities. 
% While recent STSSL \cite{ji2023spatio} addresses spatial-temporal heterogeneity through self-supervised learning, it relies on spatial-temporal graph and treats modalities as shared features. Hence, we do not consider this model as a baseline.

\subsection{Overall Performance} \label{overall_performance}
\begin{figure}[h]
  \centering
  \includegraphics[width=.9\linewidth]{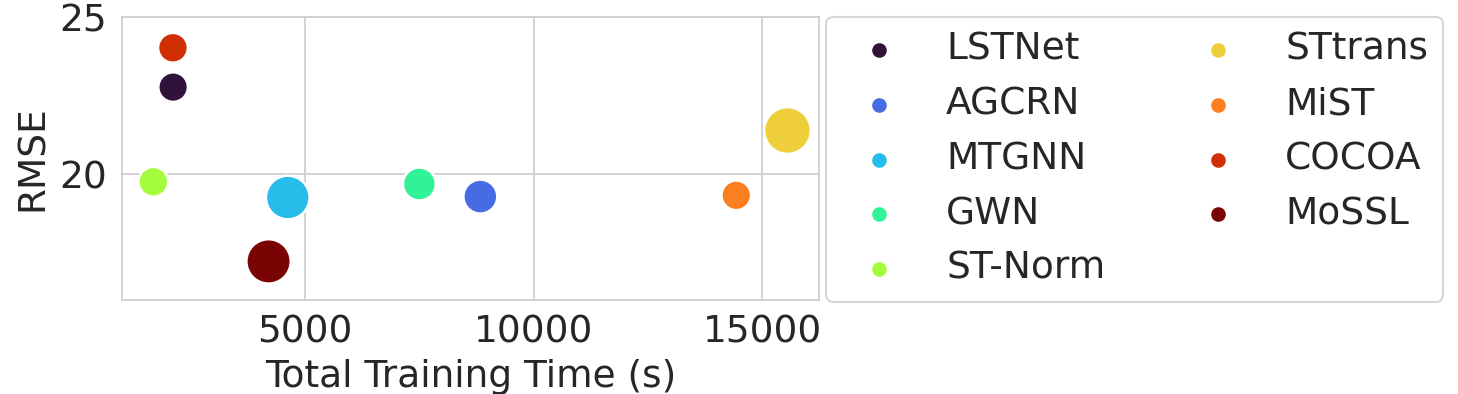}
  \caption{Efficiency study on the NYC Traffic Demand dataset.}
  \label{fig:efficiency}
\end{figure}

We compare our MoSSL with the mentioned baselines, and report the average results of five repeated experiments. 
Through Table~\ref{tab:traf}, we can find our MoSSL outperforms state-of-the-arts in all cases (modality/horizon/metric) on the NYC Traffic Demand dataset. Among the non-MoST forecasting baselines, MTGNN and GWN perform relatively well based on the WaveNet~\cite{Oord2016WaveNetAG} backbone. Models like AGCRN and ST-Norm, designed for spatio-temporal dependency modeling, exhibit a worse performance due to modality-neglected dependencies and heterogeneity, making them unsuitable for MoST forecasting tasks. COCOA shows significant performance variation among MoST forecasting baselines due to the lack of spatial modeling. While MiST maintains relatively stable performance, it falls short in fully capturing the dependencies and heterogeneity within MoST as it requires referencing modalities as auxiliary inputs to predict the target modality.
Table~\ref{tab:air} demonstrates MoSSL's superiority in almost all scenarios. The BJ Air Quality dataset only has 10 locations and less corrections, and many weather-related features are unavailable. While our MoSSL struggles with this data insufficiency, it maximizes available feature correlations. Despite the limitations, the baseline models still fall short compared to the performance of MoSSL. 
Regarding the efficiency as shown in Figure~\ref{fig:efficiency}, our MoSSL is slightly time-consuming (compared with LSTNet, ST-Norm, and COCOA) but has the smallest RMSE, indicating MoSSL strikes a balance between total training time and performance.
\subsection{Ablation Study} \label{ablation_study}
To analyze the effects of key-modules in our MoSSL framework, we perform ablation studies with several variants on the NYC Traffic Demand dataset as follows: (1) \textbf{w/o AV.} It removes the \underline{a}ugmented \underline{v}iew in the joint framework, only keeping the MSSL paradigm and using the original representation $H$ as input, with no change in sampling strategy; (2) \textbf{w/o MG.} It removes the \underline{M}ulti-Modality Data Augmentation and \underline{G}SSL, using a MoST encoder that doesn't share parameters with the original view and retaining the MSSL paradigm; (3) \textbf{w/o GSSL.} It removes the GSSL from the framework; (4) \textbf{w/o MSSL.} It removes the MSSL from the framework.
\begin{figure}[t]
  \centering
  \includegraphics[width=.95\linewidth]{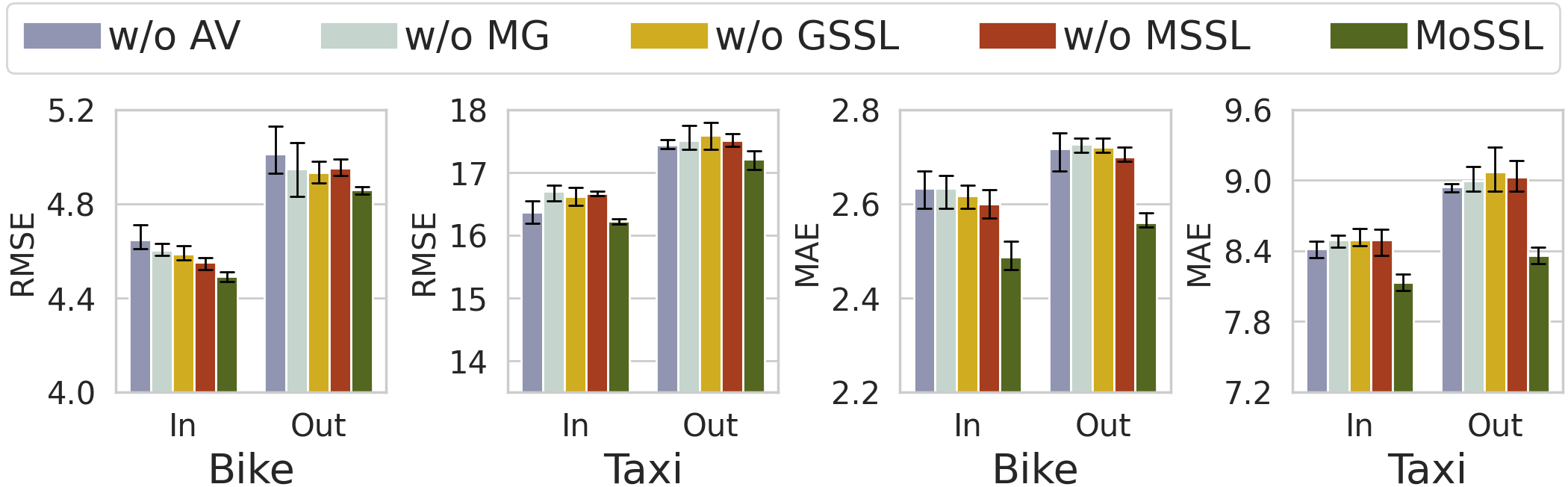}
  \caption{Ablation study of MoSSL.}
  \label{fig:ablation}
\end{figure}
The results are shown in Figure~\ref{fig:ablation}. Eliminating data augmentation, GSSL, or MSSL that relied solely on original data/representation had varying negative impacts on model performance. Overall, the complete MoSSL consistently outperforms its variations across all modalities, underscoring its holistic and indivisible nature.
\subsection{Case Study} \label{case_study}
\begin{figure}[t]
  \centering
  \includegraphics[width=.95\linewidth]{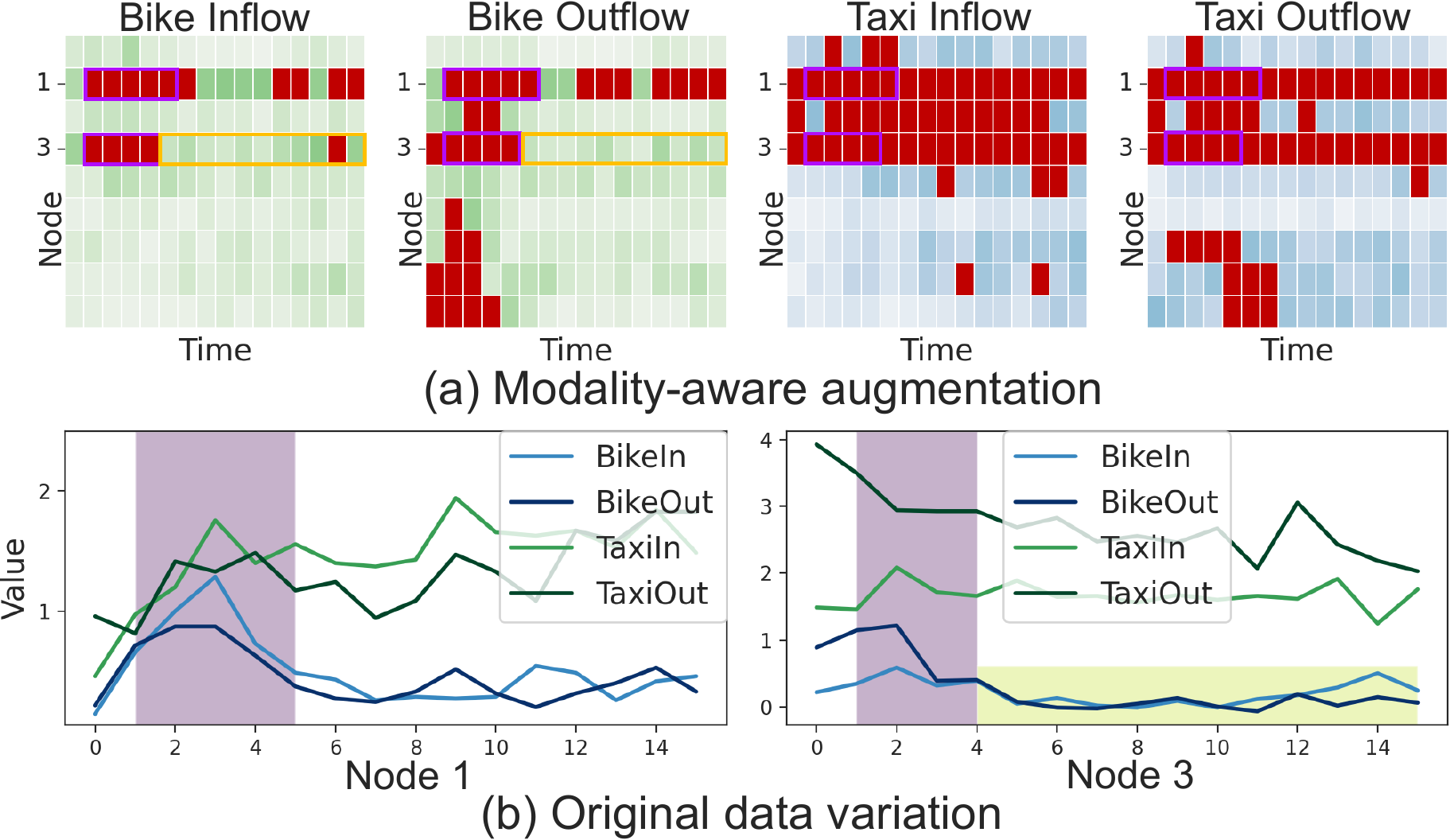}
  \caption{Case studies on modality-aware augmentation.}
  \label{fig:case_aug}
\end{figure}

\noindent \textbf{Modality Augmentation.}
We verify the Modality-aware Augmentation on the NYC Traffic Demand dataset in Figure~\ref{fig:case_aug}, where Figure~\ref{fig:case_aug}a depict heatmaps of the augmented data (red indicates masked instances), and Figure~\ref{fig:case_aug}b show the corresponding variations in node 1 and node 3 of the original data. 
Modalities exhibit distinctly different evolutions in the purple area (masked instances), while in the yellow area (retained instances), Bike Inflow and Bike Outflow display similar pattern changes. In this way, our MoSSL can adaptively eliminate biases in less correlated modality connections while preserving dependencies between modalities.

\begin{figure}[t]
  \centering
  \includegraphics[width=.95\linewidth]{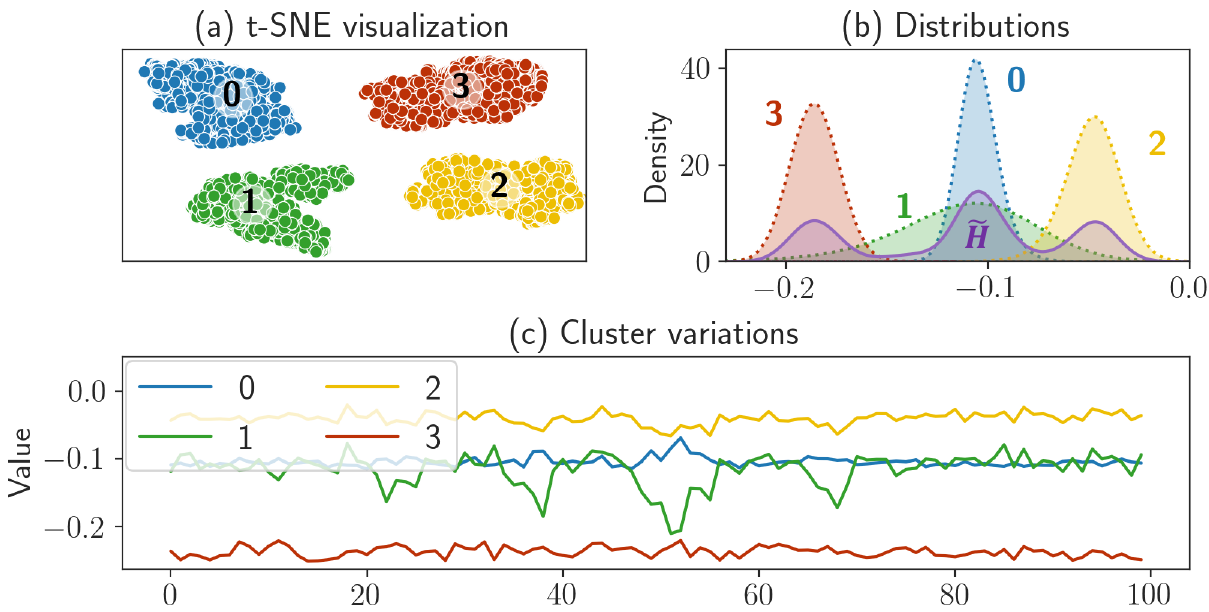}
  \caption{Case study on global heterogeneity.}
  \label{fig:case_cluster}
\end{figure}

% \noindent \textbf{Heterogeneity Disentangling.}
\noindent \textbf{Heterogeneity Disentanglement.}
We qualitatively evaluate the quality of the global heterogeneity representation learned from GSSL by visualizing them in a low-dimensional space with t-SNE, as shown in Figure~\ref{fig:case_cluster}a. We can observe that instances in the same cluster are compact, while instances in different clusters are clearly separated. Figure~\ref{fig:case_cluster}b illustrates the population distribution of representation $\Tilde{H}$ and its decomposed sub-distributions. The former, with multiple peaks, directly modeling it might cause information loss or confusion. The latter precisely quantifies heterogeneous components, enhancing the model's adaptability to MoST. We further plot the cluster variations in Figure~\ref{fig:case_cluster}c. Cluster 3 maintains relatively stable behavior over time due to its spatial dominance. Clusters 0 and 1 show intertwined evolution curves because the visualized instances stem from the same modality (Bike Inflow), and overlapping features correspond to several identical nodes. Clearly, MoSSL recognizes heterogeneities in MoST data, transferring information between similar clusters and thus aiding predictions.

\begin{figure}[t]
  \centering
  \includegraphics[width=\linewidth]{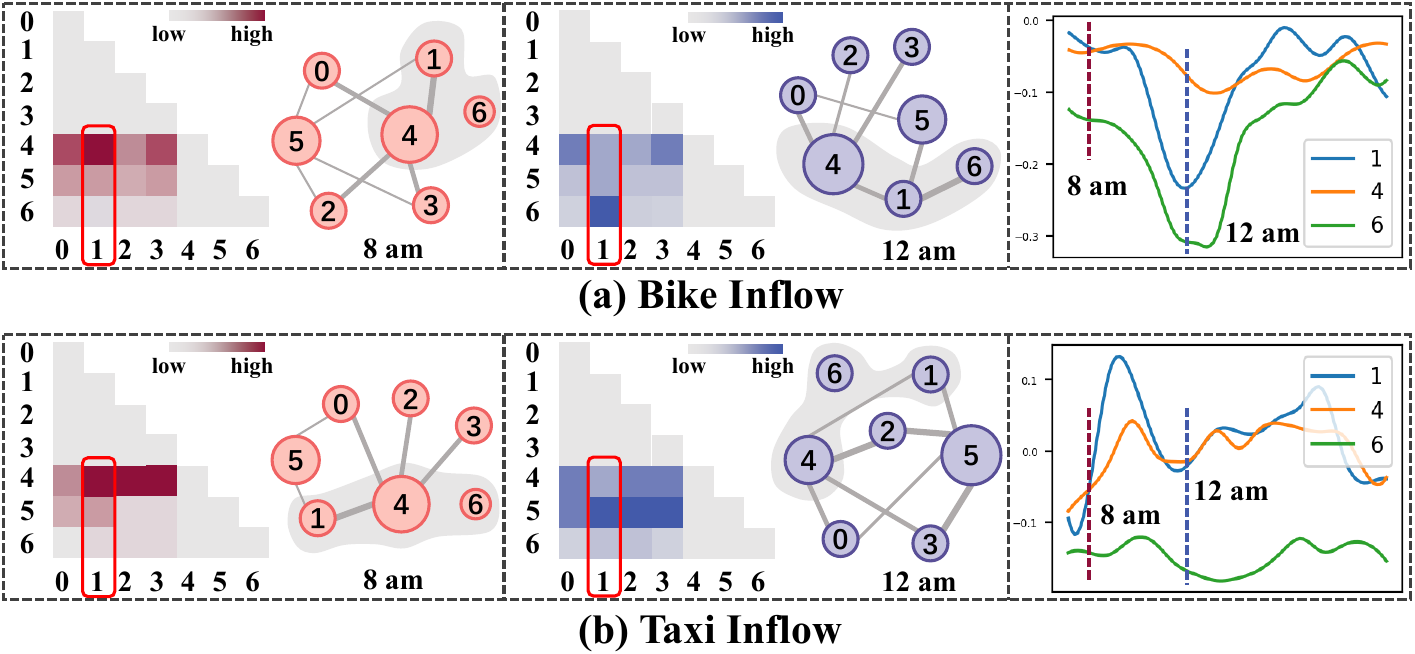}
  \caption{Case study on heterogeneity across space, time, and modality.}
  \label{fig:case_inter}
\end{figure}

% \noindent\textbf{Spatial-Temporal-Modality Heterogeneity.} 
We further investigate the explanations provided by our MoSSL in identifying crucial dynamic heterogeneity and analyzing their patterns. Figure~\ref{fig:case_inter} presents node correlation heatmaps, node relations, and corresponding evolutions for selected nodes at different modalities and times, computed using cosine similarity. In the node relation diagram, the thicker edge represents a higher correlation, and the bigger node size means a larger weighted outdegree. For Bike Inflow, as shown in Figure~\ref{fig:case_inter}a, MoSSL identifies that at 8 am, the correlation between node 1 and node 4 is stronger than with node 6, while at 12 am, the correlation weakens with node 4 and strengthens with node 6. This trend is evident in the time series diagram, where node 1 has comparable valleys with node 4 at 8 am, and a similar changing trend with node 6 at 12 am. In Taxi Inflow, as depicted in Figure~\ref{fig:case_inter}b, node 1 is more correlated with node 4 at 8 am, similar to its performance in Bike Inflow. However, at 12 am, node 1 maintains a strong correlation with node 4 and a weak correlation with node 6, contrasting with its behavior in Bike Inflow. These findings indicate that MoSSL successfully decouples heterogeneity while preserving similar patterns, avoiding blind borrowing of information from other modality/spatial/temporal trends, which facilitates predictions.

\subsection{Hyper-parameter Study} \label{hyper}
\begin{figure}[t]
  \centering
  \includegraphics[width=0.95\linewidth]{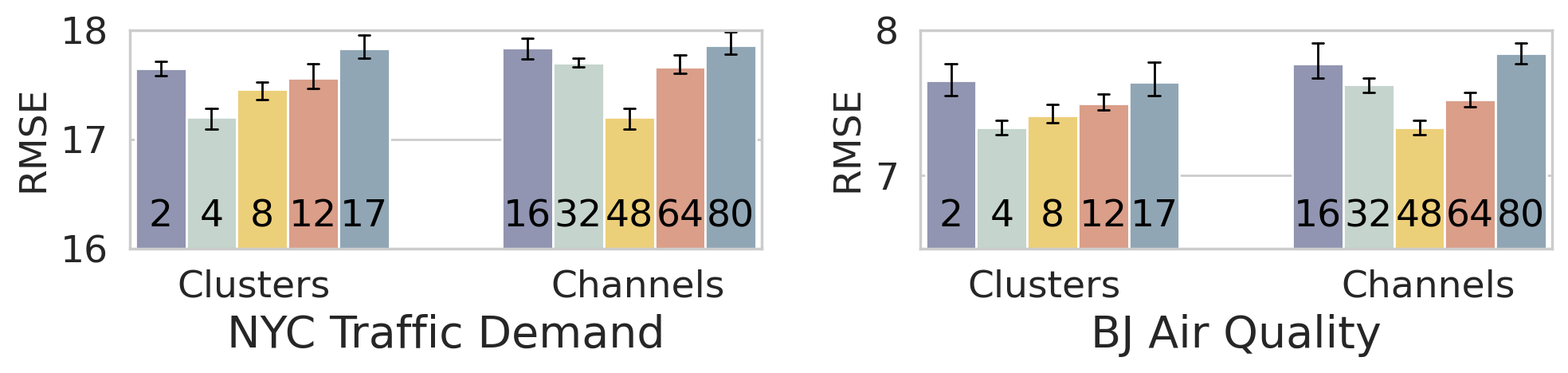}
  \caption{Hyper-parameter study of MoSSL.}
  \label{fig:hyper}
\end{figure}

We examine the impact of key hyper-parameters in Figure~\ref{fig:hyper}, i.e., the number of cluster components $K$ and the hidden channel dimension $d_z$. We observe a trend where the performance of MoSSL decreased initially and then increased with the increment of $K$ and $d_z$. This phenomenon aligns with our understanding that improper settings of cluster components and channel dimensions can lead to underfitting or overfitting of the model. The model achieved comparable performance when $K$ and $d_z$ are set to 4 and 48, respectively.

\section{Related Work}
Many studies have been devoted to developing deep learning-based time series forecasting. As the canonical cases of spatio-temporal forecasting, graph neural network is developed to model the spatial relation \cite{li2017diffusion,oreshkin2021fc}; attention mechanisms \cite{vaswani2017attention} and its variants include spatial attention \cite{fang2019gstnet,zheng2020gman}, temporal attention \cite{wu2021autoformer,zheng2020gman}, and adaptive embedding \cite{liu2023spatio}; convolution operators are utilized in spatial convolution \cite{deng2022graph,guo2021hierarchical}, temporal convolution \cite{wu2020connecting,wu2019graph}, spatio-temporal convolution \cite{guo2019deep,yang2021space} and adaptive convolution \cite{pan2019urban}. \cite{jiang2021dl} establishes a benchmark for prevalent deep learning models. \cite{zhu2021mixseq} is proposed for micro-cluster analysis and prediction in macro time series. \cite{fan2023dish} models the inter- and intra-space distribution drift and \cite{fan2022depts} learns the periodicity of the time series. Traditional spatio-temporal modeling approaches might not entirely fit MoST because of the dynamic and heterogeneity arising from the additional modality domain. 
Some recent methods extend spatio-temporal modeling by considering additional domains (i.e., modality or categorical domain) \cite{ye2019co,huang2019mist,han2021dynamic,wang2021spatio} or data sources \cite{ijcai2023p231,jiang2023learning,deng2023tts}, extracting insights from different modalities as auxiliary inputs to leverage similar historical patterns. However, they struggle to fully explore heterogeneity or interactions from different modalities. \cite{ji2023spatio,10.1145/3550316} model time series with self-supervised learning but lack modality sensitivity (shared information among modalities) and spatial sensitivity (ignoring spatial characteristics). Our approach stands out by integrating multi-modality data augmentation and two self-supervised learning paradigms, explicitly capturing similar patterns while quantifying heterogeneity to better adapt to MoST data.
\section{Conclusion}
In this study, we propose a self-supervised learning framework MoSSL for multi-modality spatio-temporal forecasting. Specifically, we integrate Multi-Modality Data Augmentation and two self-supervised learning paradigms, i.e., GSSL for comprehending diverse pattern changes in space, time, and modality, and MSSL for strengthening the learned representations of inter- and intra-modality features. Extensive experiments on two real-world MoST datasets verify the superiority of MoSSL. In future work, we will refine the model to be more lightweight and universally applicable.
\section*{Acknowledgments}
This work was partially  supported by the grants of National Key Research and Development Project (2021YFB1714400) of China and Jilin Provincial International Cooperation Key Laboratory for Super Smart City.

\bibliographystyle{named}
\bibliography{ijcai24}
\end{document}